\documentclass[runningheads]{llncs}

\usepackage[utf8]{inputenc}
\usepackage{caption}
\usepackage{subcaption}
\usepackage{graphicx}
\usepackage{lscape}
\usepackage{booktabs}
\usepackage{multirow}
\usepackage{amssymb}

\DeclareUnicodeCharacter{FB01}{fi}

\usepackage{array}
\newcolumntype{L}[1]{>{\raggedright\let\newline\\\arraybackslash\hspace{0pt}}m{#1}}
\newcolumntype{C}[1]{>{\centering\let\newline\\\arraybackslash\hspace{0pt}}m{#1}}
\newcolumntype{R}[1]{>{\raggedleft\let\newline\\\arraybackslash\hspace{0pt}}m{#1}}

\begin{document}
\title{Spoken Language Identification using ConvNets}
\author{Sarthak\inst{1} \and
Shikhar Shukla\inst{2} \and
Govind Mittal\inst{3}}
\authorrunning{Sarthak et al.}
\institute{Analytics Quotient, Bangalore, India \\
\email{sarthak.sfc@gmail.com, sarthak.j@aqinsights.com}\\ \and
Samsung R\&D Institute India-Bangalore, Bangalore, India\\
\email{shikhar.00778@gmail.com, shikhar.0077@samsung.com} \and
Birla Institute of Technology \& Science, Pilani, Rajasthan, India \\
\email{f2014530@pilani.bits-pilani.ac.in}
}
\maketitle              % typeset the header of the contribution
\begin{abstract}
Language Identification (LI) is an important first step in several speech processing systems. With a growing number of voice-based assistants, speech LI has emerged as a widely researched field. To approach the problem of identifying languages, we can either adopt an implicit approach where only the speech for a language is present or an explicit one where text is available with its corresponding transcript. This paper focuses on an implicit approach due to the absence of transcriptive data. This paper benchmarks existing models and proposes a new attention based model for language identification which uses log-Mel spectrogram images as input. We also present the effectiveness of raw waveforms as features to neural network models for LI tasks. For training and evaluation of models, we classified six languages (English, French, German, Spanish, Russian and Italian) with an accuracy of 95.4\% and four languages (English, French, German, Spanish) with an accuracy of 96.3\% obtained from the VoxForge dataset. This approach can further be scaled to incorporate more languages.

\end{abstract}
%
%
%g
\section{Introduction}
Language Identification (LI) is a problem which involves classifying the language being spoken by a speaker. LI systems can be used in call centers to route international calls to an operator who is fluent in that identified language \cite{kumar2010spoken}. In speech-based assistants, LI acts as the first step which chooses the corresponding grammar from a list of available languages for its further semantic analysis \cite{bartz2017language}. It can also be used in multi-lingual voice-controlled information retrieval systems, for example, Apple Siri and Amazon Alexa.

Over the years, studies have utilized many prosodic and acoustic features to construct machine learning models for LI systems \cite{obuchi2005language}. Every language is composed of \emph{phonemes}, which are distinct unit of sounds in that language, such as \emph{b} of black and \emph{g} of green. Several prosodic and acoustic features are based on phonemes, which become the underlying features on whom the performance of the statistical model depends \cite{1659993,article}. If two languages have many overlapping phonemes, then identifying them  becomes a challenging task for a classifier. For example, the word \emph{cat} in English, \emph{kat} in Dutch, \emph{katze} in German have different consonants but when used in a speech they all would sound quite similar. 

Due to such drawbacks several studies have switched over to using Deep Neural Networks (DNNs) to harness their novel auto-extraction techniques \cite{bartz2017language,revay2019multiclass}. This work follows an implicit approach for identifying six languages with overlapping phonemes on the VoxForge \cite{VoxForge.org} dataset and achieves 95.4\% overall accuracy. 

In previous studies \cite{bartz2017language,montavon2009deep,revay2019multiclass}, authors use log-Mel spectrum of a raw audio as inputs to their models. One of our contributions is to enhance the performance of this approach by utilising recent techniques like Mixup augmentation of inputs and exploring the effectiveness of \emph{Attention} mechanism in enhancing performance of neural network. As log-Mel spectrum needs to be computed for each raw audio input and processing time for generating log-Mel spectrum increases linearly with length of audio, this acts as a bottleneck for these models. Hence, we propose the use of raw audio waveforms as inputs to deep neural network which boosts performance by avoiding additional overhead of computing log-Mel spectrum for each audio. Our 1D-ConvNet architecture auto-extracts and classifies features from this raw audio input.
 
The structure of the work is as follows. In Section 2 we discuss about the previous related studies in this field. The model architecture for both the raw waveforms and log-Mel spectrogram images is discussed in Section 3 along with the a discussion on hyperparameter space exploration. In Section 4 we present the experimental results. Finally, in Section 5 we discuss the conclusions drawn from the experiment and future work.

\section{Related Work}

Extraction of language dependent features like prosody and phonemes was a popular approach to classify spoken languages \cite{zissman1996comparison,martinez2011language,ferrer2010comparison}. Following their success in speaker verification systems, i-vectors have also been used as features in various classification networks. These approaches required significant domain knowledge \cite{dehak2011language,martinez2011language}. Nowadays most of the attempts on spoken language identification rely on neural networks for meaningful feature extraction and classification \cite{lopez2014automatic,ganapathy2014robust}.

Revay et al. \cite{revay2019multiclass} used the ResNet50 \cite{DBLP:conf/cvpr/HeZRS16} architecture for classifying languages by generating the log-Mel spectra of each raw audio. The model uses a cyclic learning rate where learning rate increases and then decreases linearly. Maximum learning rate for a cycle is set by finding the optimal learning rate using \emph{fastai} \cite{FastAI} library. The model classified six languages -- English, French, Spanish, Russian, Italian and German -- and achieving an accuracy of 89.0\%.

Gazeau et al. \cite{gazeau2018automatic} in his research showed how Neural Networks, Support Vector Machine and Hidden Markov Model (HMM) can be used to identify French, English, Spanish and German. Dataset was prepared using voice samples from Youtube News \cite{Youtube}and VoxForge \cite{VoxForge.org} datasets. Hidden Markov models convert speech into a sequence of vectors, was used to capture temporal features in speech. HMMs trained on VoxForge \cite{VoxForge.org} dataset performed best in comparison to other models proposed by him on same VoxForge dataset. They reported an accuracy of 70.0\%.

Bartz et al. \cite{bartz2017language} proposed two different hybrid Convolutional Recurrent Neural Networks for language identification. They proposed a new architecture for extracting spatial features from log-Mel spectra of raw audio using CNNs and then using RNNs for capturing temporal features to identify the language. This model achieved an accuracy of 91.0\% on Youtube News Dataset \cite{Youtube}. In their second architecture they used the Inception-v3 \cite{szegedy2016rethinking} architecture to extract spatial features which were then used as input for bi-directional LSTMs to predict the language accurately. This model achieved an accuracy of 96.0\% on four languages which were English, German, French and Spanish. They also trained their CNN model (obtained after removing RNN from CRNN model) and the Inception-v3 on their dataset. However they were not able to achieve better results achieving and reported 90\% and 95\% accuracies, respectively. 

Kumar et al. \cite{kumar2010spoken} used Mel-frequency cepstral coefficients (MFCC), Perceptual linear prediction coefficients (PLP), Bark Frequency Cepstral Coefficients (BFCC) and Revised Perceptual Linear Prediction Coefficients (RPLP) as features for language identification. BFCC and RPLP are hybrid features derived using MFCC and PLP. They used two different models based on Vector Quantization (VQ) with Dynamic Time Warping (DTW) and Gaussian Mixture Model (GMM) for classification. These classification models were trained with different features. The authors were able to show that these models worked better with hybrid features (BFCC and RPLP) as compared to conventional features (MFCC and PLP). GMM combined with RPLP features gave the most promising results and achieved an accuracy of 88.8\% on ten languages. They designed their own dataset comprising of ten languages being Dutch, English, French, German, Italian, Russian, Spanish, Hindi, Telegu, and Bengali. 

Montavon \cite{montavon2009deep} generated Mel spectrogram as features for a time-delay neural network (TDNN). This network had two-dimensional convolutional layers for feature extraction. An elaborate analysis of how deep architectures outperform their shallow counterparts is presented in this reseacrch. The difficulties in classifying perceptually similar languages like German and English were also put forward in this work. It is mentioned that the proposed approach is less robust to new speakers present in the test dataset. This method was able to achieve an accuracy of 91.2\% on dataset comprising of 3 languages -- English, French and German.

In Table~\ref{tab:prev}, we summarize the quantitative results of the above previous studies. It includes the model basis, feature description, languages classified and the used dataset along with accuracy obtained. The table also lists the overall results of our proposed models (at the top). The languages used by various authors along with their acronyms are English (En), Spanish (Es), French (Fr), German (De), Russian (Ru), Italian (It), Bengali (Ben), Hindi (Hi) and Telegu (Tel).

\begin{landscape}
\begin{table*}[]
\centering
\caption{Quantitative Review of Previous Studies along with our Results.}
\label{tab:prev}
% \resizebox{\textwidth}{!}{%
\begin{tabular}{L{1cm}L{2.5cm}L{1.5cm}L{2.2cm}L{1cm}L{8.9cm}L{1cm}}
\toprule
\textbf{Year} & \textbf{Model basis} & \textbf{Features} & \textbf{Languages} & \textbf{Acc.} & \textbf{Remarks} & \textbf{Ref.} \\ \midrule
2019 & 1D ConvNet & Raw Audio & En, Fr, De, Es, Ru, It & 93.7$^1$ & Evaulation of our 1D ConvNet model with mixup for six languages. & self \\ \midrule
2019 & 2D ConvNet & log-Mel & En, Fr, De, Es, Ru, It & 95.4$^1$ & Evaulation of our 2D ConvNet model with mixup for six languages. & self \\ \midrule
2019 & 2D ConvNet-Bi-directional GRU-Attention & log-Mel & En, Fr, De, Es, Ru, It & 95.0$^1$ &
 Result after tuning the hyperparameters of our cnn-bi-directional GRU-attention model and applying mixup & self \\ \midrule
2019 & 2D ConvNet & log-Mel & En, Fr, De, Es & 96.3$^1$ & Our evaluation of 2D ConvNet model for  four languages. & self \\ \midrule
2019 & ResNet50 & log-Mel & En, Fr, De, Es, Ru, It & 89.0$^1$ & Uses a pretrained ResNet50 architecture and cyclic learner to identify the language. & \cite{revay2019multiclass} \\ \midrule
2018 & SVM-HMM model & not defined & En, Fr, Es, De & 70.0$^1$ & HMMs were used to encode speech into sequences of vectors which were then fed into a neural network. & \cite{gazeau2018automatic} \\ \midrule
2017 & Inceptionv3 CRNN & log-Mel & En, Fr, De, Es & 96.0$^2$ & Used Inception-v3 model followed by bi-directional LSTMs to extract convolutional and temporal features. & \cite{bartz2017language} \\ \midrule
2017 & CRNN & log-Mel & En, Fr, De, Es & 91.0$^2$ & A new architecture is used to extract spatial features by using CNNs and temporal features using RNNs. & \cite{bartz2017language} \\ \midrule
2010 & Gaussian Mixture Models & Perceptual Linear Prediction & Dut, En, Fr, De, It, Ru, Es, Ben, Hi and Tel & 88.8$^3$ & Used Gaussian mixture models coupled with RPLP features, which were prepared using MFCC and PLP. & \cite{kumar2010spoken} \\ \midrule
2009 & CNN-TDNN & log-Mel & En, Fr , De & 91.2$^1$ & Used a time delay neural network with SGD was used to identify language using log-Mel images as input. & \cite{montavon2009deep} \\ \bottomrule
\multicolumn{6}{l}{Dataset: $^1$ - VoxForge \cite{VoxForge.org}; $^2$ - Youtube News \cite{Youtube}, $^3$ - Private} \\ 
\end{tabular}%
% }
\end{table*}
\end{landscape}

\section{Proposed Method}

\subsection{Motivations}

Several state-of-the-art results on various audio classification tasks have been obtained by using log-Mel spectrograms of raw audio, as features \cite{xu2019general}. Convolutional Neural Networks have demonstrated an excellent performance gain in classification of these features \cite{xu2017unsupervised,hershey2017cnn} against other machine learning techniques. It has been shown that using \emph{attention} layers with ConvNets further enhanced their performance \cite{lee2017raw}. This motivated us to develop a CNN-based architecture with \emph{attention} since this approach hasn’t been applied to the task of language identification before.

Recently, using raw audio waveform as features to neural networks has become a popular approach in audio classification \cite{weireport,lee2017raw}. Raw waveforms have several artifacts which are not effectively captured by various conventional feature extraction techniques like Mel Frequency Cepstral Coefficients (MFCC), Constant Q Transform (CQT), Fast Fourier Transform (FFT), etc.

Audio files are a sequence of spoken words, hence they have temporal features too.A CNN is better at capturing spatial features only and RNNs are better at capturing temporal features as demonstrated by Bartz et al. \cite{bartz2017language} using audio files. Therefore, we combined both of these to make a CRNN model.

We propose three types of models to tackle the problem with different approaches, discussed as follows. 

\subsection{Description of Features}

As an average human's voice is around 300 Hz and according to Nyquist-Shannon sampling theorem all the useful frequencies (0-300 Hz) are preserved with sampling at 8 kHz, therefore, we sampled \emph{raw audio} files from all six languages at 8 kHz

The average length of audio files in this dataset was about \emph{10.4 seconds} and standard deviation was \emph{2.3 seconds}. For our experiments, the audio length was set to 10 seconds. If the audio files were shorter than 10 second, then the data was repeated and concatenated. If audio files were longer, then the data was truncated. 

\subsection{Model Description}

We applied the following design principles to all our models:

\begin{itemize}
	\item \emph{Every convolutional layer is always followed by an appropriate max pooling layer}. This helps in containing the explosion of parameters and keeps the model small and nimble.
	\item \emph{Convolutional blocks} are defined as an individual block with multiple pairs of one convolutional layer and one max pooling layer. \emph{Each convolutional block is preceded or succeded by a convolutional layer.}
    \item \emph{Batch Normalization and Rectified linear unit activations were applied after each convolutional layer}. Batch Normalization helps speed up convergence during training of a neural network.
	\item Model \emph{ends with a dense layer} which acts the final output layer.
\end{itemize}

\subsection{Model Details: 1D ConvNet}
\label{sec:1dcnn}
As the sampling rate is 8 kHz and audio length is 10 s, hence the input is \emph{raw audio} to the models with input size of (batch size, 1, 80000). In Table~\ref{tab:cnn1d}, we present a detailed layer-by-layer illustration of the model along with its hyperparameter.

%1dcnn
\begin{table}[h!]
 \centering
 \setlength{\belowcaptionskip}{-10pt}
 \caption{Architecture of the 1D-ConvNet model}
 \label{tab:cnn1d}
 \resizebox{\textwidth}{!}{%
 \begin{tabular}{|l|l|l|l|}
 \hline
 \textbf{Layer Name} & \textbf{\begin{tabular}[c]{@{}l@{}}\# filters / kernel /\\  stride\end{tabular}} & \textbf{output} & \textbf{\# of parameters} \\ \hline
 Conv1 & (128, 3, 3) & (128, 26664) & 384 \\ \hline
 \begin{tabular}[c]{@{}l@{}}(Convolutional Block 1)\\ Conv1D\\ MaxPool1D\\ Conv1D\\ MaxPool1D\end{tabular} & \begin{tabular}[c]{@{}l@{}}(128, 3, 1)\\ (3, 3)\\ (128, 3, 1)\\ (3, 3)\end{tabular} & \begin{tabular}[c]{@{}l@{}}(128, 26658)\\ (128, 8880)\\ (128, 8880)\\ (128, 2960)\end{tabular} & \begin{tabular}[c]{@{}l@{}}49152\\ \\ 49,152\end{tabular} \\ \hline
 \begin{tabular}[c]{@{}l@{}}Conv1D\\ MaxPool1D\end{tabular} & \begin{tabular}[c]{@{}l@{}}(256, 3, 1)\\ (3, 3)\end{tabular} & \begin{tabular}[c]{@{}l@{}}(256, 2954)\\ (256, 984)\end{tabular} & 98,304 \\ \hline
 \begin{tabular}[c]{@{}l@{}}(Convolutional Block 2)\\ Conv1D\\ MaxPool1D\end{tabular} & \begin{tabular}[c]{@{}l@{}}(256, 3, 1)\\ (3, 3)\end{tabular} & \begin{tabular}[c]{@{}l@{}}(256, 978)\\ (256, 326)\end{tabular} & 196,608 \\ \hline
 \begin{tabular}[c]{@{}l@{}}Conv1D\\ MaxPool1D\end{tabular} & \begin{tabular}[c]{@{}l@{}}(512, 3, 1)\\ (106, 3)\end{tabular} & \begin{tabular}[c]{@{}l@{}}(512, 320)\\ (512, 1)\end{tabular} & 393,216 \\ \hline
 Dense Layer & (512, 6) & (6) & 3,072 \\ \hline
 \end{tabular}%
 }
\vspace{-10mm}
\end{table}
 
\setlength{\belowcaptionskip}{-10pt}
\subsubsection{Hyperparameter Optimization:}
\begin{figure}[t]
 \centering
 \resizebox{\textwidth}{!}{%
 \includegraphics[width=\textwidth]{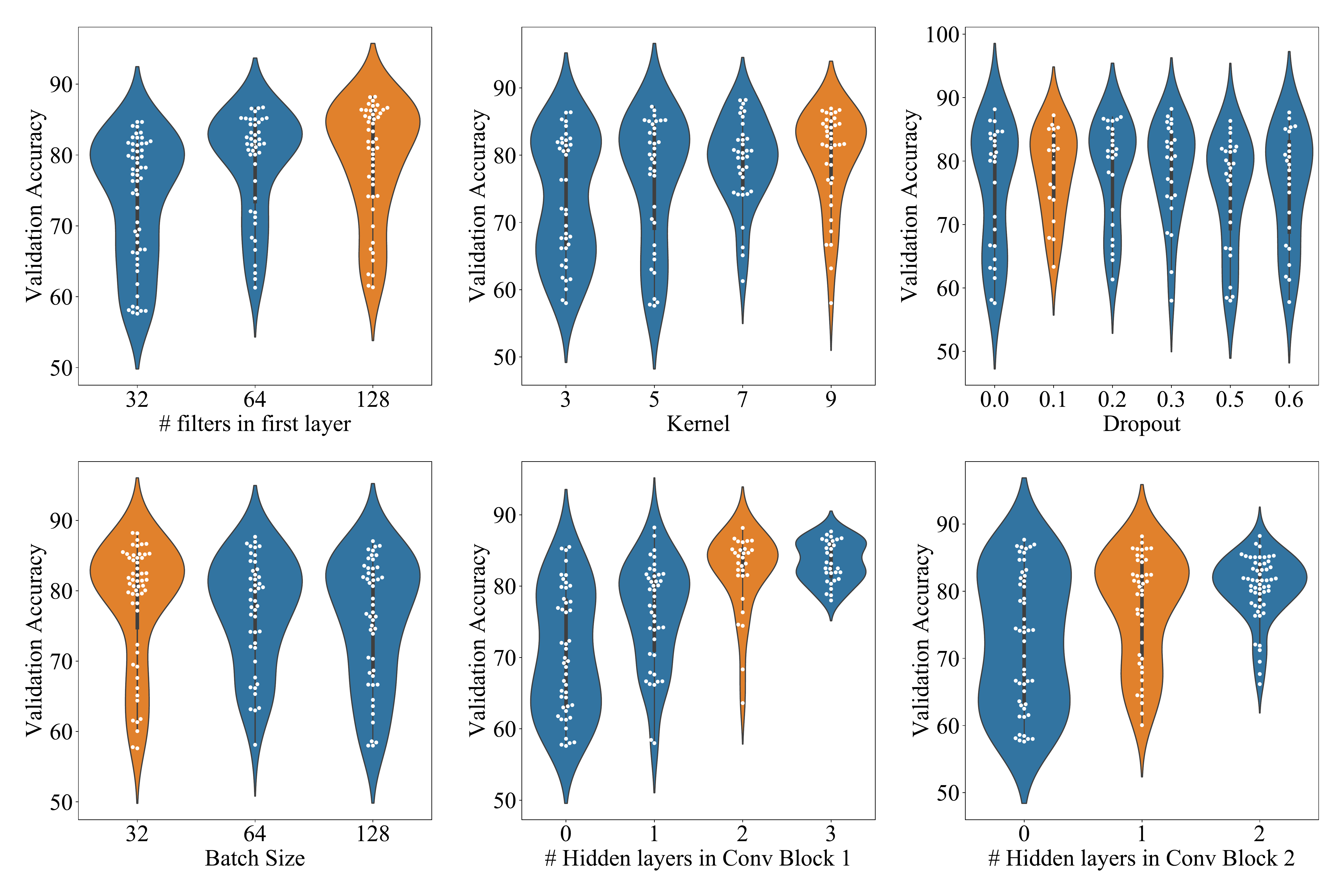}%
 }
 \setlength{\belowcaptionskip}{-20pt}
 \caption {Effect of hyperparameter variation of the hyperparameter on the  classification accuracy for the case of 1D-ConvNet. Orange colored violin plots show the most favored choice of the hyperparameter and blue shows otherwise. One dot represents one sample.}
 \label{fig:1dcnn}
\end{figure}

Tuning hyperparameters is a cumbersome process as the hyperparamter space expands exponentially with the number of parameters, therefore efficient exploration is needed for any feasible study. We used the \emph{random search} algorithm supported by \emph{Hyperopt} \cite{bergstra2013making} library to randomly search for an optimal set of hyperparameters from a given parameter space. In Fig.~\ref{fig:1dcnn}, various hyperparameters we considered are plotted against the validation accuracy as violin plots. Our observations for each hyperparameter are summarized below:

\emph{Number of filters in first layer}: We observe that having 128 filters gives better results as compared to other filter values of 32 and 64 in the first layer. A higher number of filters in the first layer of network is able to preserve most of the characteristics of input.

\emph{Kernel Size}: We varied the receptive fields of convolutional layers by choosing the kernel size from among the set of \{3, 5, 7, 9\}. We observe that a kernel size of 9 gives better accuracy at the cost of increased computation time and larger number of parameters. A large kernel size is able to capture longer patterns in its input due to bigger receptive power which results in an improved accuracy.

\emph{Dropout}: Dropout randomly turns-off (sets to 0) various individual nodes during training of the network. In a deep CNN it is important that nodes do not develop a co-dependency amongst each other during training in order to prevent overfitting on training data \cite{srivastava2014dropout}. Dropout rate of $0.1$ works well for our model. When using a higher dropout rate the network is not able to capture the patterns in training dataset.

\emph{Batch Size}: We chose batch sizes from amongst the set \{32, 64, 128\}. There is more noise while calculating error in a smaller batch size as compared to a larger one. This tends to have a regularizing effect during training of the network and hence gives better results. Thus, batch size of 32 works best for the model.

\emph{Layers in Convolutional block 1 and 2}: We varied the number of layers in both the convolutional blocks. If the number of layers is low, then the network does not have enough depth to capture patterns in the data whereas having large number of layers leads to overfitting on the data. In our network, two layers in the first block and one layer in the second block give optimal results.

% 1d convNet hyper

\subsection{Model Details: 2D ConvNet with Attention and bi-directional GRU}
\label{sec:2dgru}
Log-Mel spectrogram is the most commonly used method for converting audio into the image domain. The audio data was again sampled at 8 kHz. The input to this model was the log-Mel spectra. We generated log-Mel spectrogram using the \emph{LibROSA} \cite{Librosa} library. In Table~\ref{tab:cnn2d}, we present a detailed layer-by-layer illustration of the model along with its hyperparameter.

\subsubsection{} We took some specific design choices for this model, which are as follows:
\begin{itemize}
 \item We added \emph{residual connections} with each convolutional layer. Residual connections in a way makes the model selective of the contributing layers, determines the optimal number of layers required for training and solves the problem of vanishing gradients. Residual connections or skip connections skip training of those layers that do not contribute much in the overall outcome of model.

%2dcnn
\begin{table}[]
\centering
 \setlength{\belowcaptionskip}{-5pt}
 \caption{Architecture of the 2D-ConvNet model}
 
 \label{tab:cnn2d}
\resizebox{\textwidth}{!}{%
\begin{tabular}{|l|l|l|l|}
\hline
Layer Name & Output features & Number of filters / stride / padding & No. of parameters \\\hline 
\begin{tabular}[c]{@{}l@{}}(ConvBlock 1)\\ Conv2D\\ Conv2D\\ AvgPool2D\end{tabular} & \begin{tabular}[c]{@{}l@{}}(64, 128, 128)\\ (64, 128, 128)\\ (64, 64, 64)\end{tabular} & \begin{tabular}[c]{@{}l@{}}(3, 3) / (1, 1) / (1, 1)\\ (3, 3) / (1, 1) / (1, 1)\end{tabular} & \begin{tabular}[c]{@{}l@{}}1,728\\ 36,864\end{tabular} \\ \hline
\begin{tabular}[c]{@{}l@{}}(ConvBlock 2)\\ Conv2D\\ Conv2D\\ AvgPool2D\end{tabular} & \begin{tabular}[c]{@{}l@{}}(128, 64, 64)\\ (128, 64, 64)\\ (128, 32, 32)\end{tabular} & \begin{tabular}[c]{@{}l@{}}(3, 3) / (1, 1) / (1, 1)\\ (3, 3) / (1, 1) / (1, 1)\end{tabular} & \begin{tabular}[c]{@{}l@{}}73,728\\ 147,456\end{tabular} \\ \hline
\begin{tabular}[c]{@{}l@{}}(ConvBlock 3)\\ Conv2D\\ Conv2D\\ AvgPool2D\end{tabular} & \begin{tabular}[c]{@{}l@{}}(256, 32, 32)\\ (256, 32, 32)\\ (256, 16, 16)\end{tabular} & \begin{tabular}[c]{@{}l@{}}(3, 3) / (1, 1) / (1, 1)\\ (3, 3) / (1, 1) / (1, 1)\end{tabular} & \begin{tabular}[c]{@{}l@{}}294,912\\ 589,824\end{tabular} \\ \hline
\begin{tabular}[c]{@{}l@{}}(ConvBlock 4)\\ Conv2D\\ Conv2D\\ AvgPool2D\end{tabular} & \begin{tabular}[c]{@{}l@{}}(512, 16, 16)\\ (512, 16, 16)\\ (512, 8, 8)\end{tabular} & \begin{tabular}[c]{@{}l@{}}(3, 3) / (1, 1) / (1, 1)\\ (3, 3) / (1, 1) / (1, 1)\end{tabular} & \begin{tabular}[c]{@{}l@{}}1,179,648\\ 235,929\end{tabular} \\ \hline
\begin{tabular}[c]{@{}l@{}}Bi-directional GRU\\ Embedding Layer\end{tabular} & \begin{tabular}[c]{@{}l@{}}(8, 1536)\\ (8, 768)\end{tabular} &  & \begin{tabular}[c]{@{}l@{}}1,769,472\\ 1,179,648\end{tabular} \\ \hline
\begin{tabular}[c]{@{}l@{}}(Sequential Block)\\ Dropout (0.2)\\ Linear\\ Dropout (0.1)\\ Linear\end{tabular} & \begin{tabular}[c]{@{}l@{}}(256)\\ \\ (6)\end{tabular} &  & \begin{tabular}[c]{@{}l@{}}131,072\\ \\ 1,536\end{tabular} \\ \hline
\end{tabular}%
}
\vspace{1mm}
\end{table}

 \item We added \emph{spatial attention} \cite{chen2017sca} networks to help the model in focusing on specific regions or areas in an image. Spatial attention aids learning irrespective of transformations, scaling and rotation done on the input images making the model more robust and helping it to achieve better results.
 
 \item We added \emph{Channel Attention} networks so as to help the model to find interdependencies among color channels of log-Mel spectra. It adaptively assigns importance to each color channel in a deep convolutional multi-channel network. In our model we apply channel and spatial attention just before feeding the input into bi-directional GRU. This helps the model to focus on selected regions and at the same time find patterns among channels to better determine the language.
\end{itemize}
 
% 2d convNet hyper
\begin{figure}
 \centering
 \resizebox{0.95\textwidth}{!}{%
 \includegraphics[width=\textwidth]{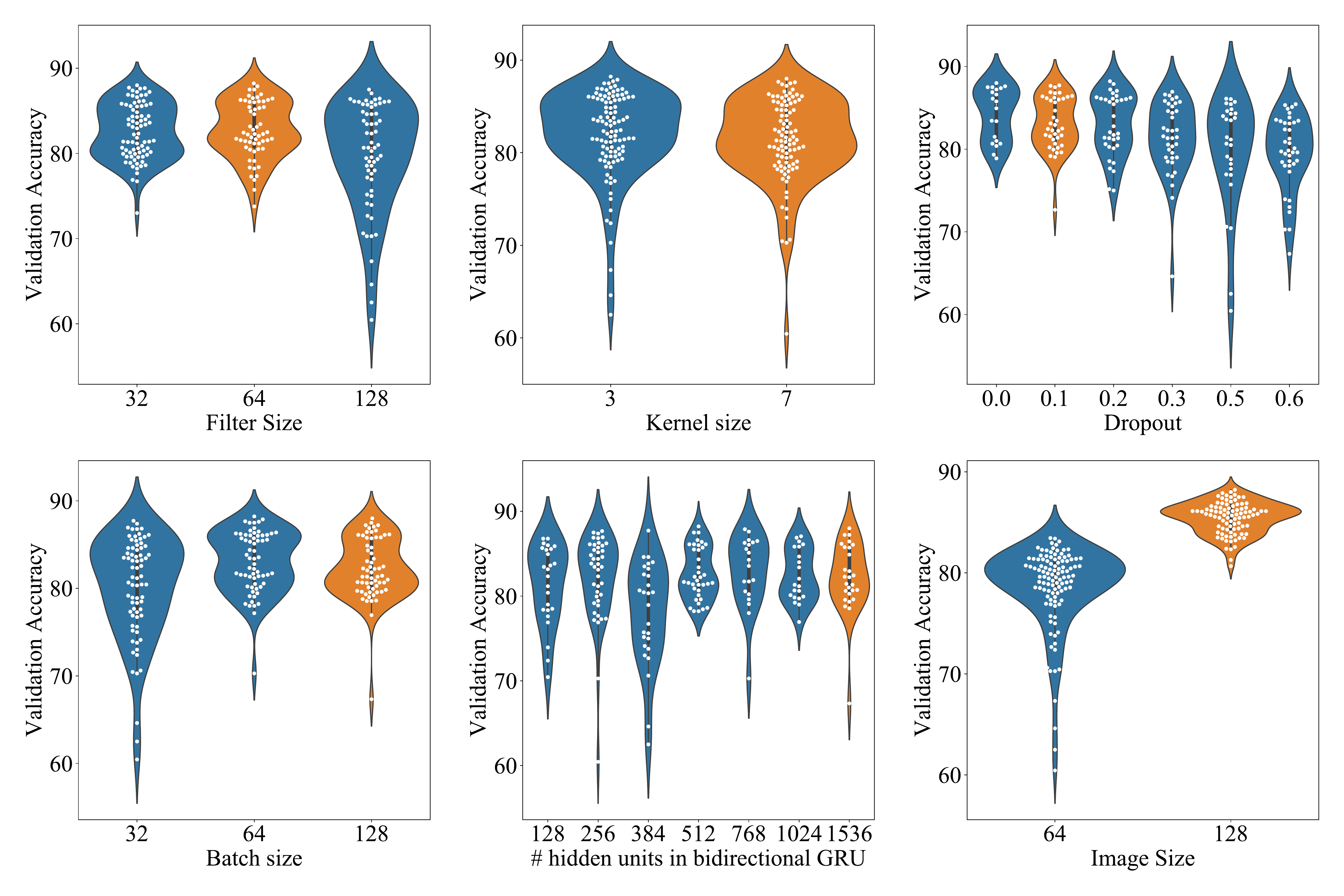}
 }
 \setlength{\belowcaptionskip}{-33pt}
 \caption {Effect of hyperparameter variation of the six selected hyperparameter on the  classification accuracy for the case of 2D-ConvNet. Orange colored violin plots show the most favored choice of the hyperparameter and blue shows otherwise. One dot represents one sample.}
 \label{fig:2dcnn}
\end{figure}

 \subsubsection{Hyperparameter Optimization:} We used the \emph{random search} algorithm supported by \emph{Hyperopt} \cite{bergstra2013making} library to randomly search for an optimal set of hyperparameters from a given parameter space. In Fig.~\ref{fig:2dcnn} ,various hyperparameters we tuned are plotted against the validation accuracy. Our observations for each hyperparameter are summarized below:

 \emph{Filter Size}: 64 filters in the first layer of network can preserve most of the characteristics of input, but increasing it to 128 is inefficient as overfitting occurs.

 \emph{Kernel Size}: There is a trade-off between kernel size and capturing complex non-linear features. Using a small kernel size will require more layers to capture features whereas using a large kernel size will require less layers. Large kernels capture simple non-linear features whereas using a smaller kernel will help us capture more complex non-linear features. However, with more layers, backpropagation necessitates the need for a large memory. We experimented with large kernel size and gradually increased the layers in order to capture more complex features. The results are not conclusive and thus we chose kernel size of 7 against 3.

 \emph{Dropout}: Dropout rate of 0.1 works well for our data. When using a higher dropout rate the network is not able to capture the patterns in training dataset.

 \emph{Batch Size}: There is always a trade-off between batch size and getting accurate gradients. Using a large batch size helps the model to get more accurate gradients since the model tries to optimize gradients over a large set of images. We found that using a batch size of 128 helped the model to train faster and get better results than using a batch size less than 128. 

 \emph{Number of hidden units in bi-directional GRU}: Varying the number of hidden units and layers in GRU helps the model to capture temporal features which can play a significant role in identifying the language correctly. The optimal number of hidden units and layers depends on the complexity of the dataset. Using less number of hidden units may capture less features whereas using large number of hidden units may be computationally expensive. In our case we found that using 1536 hidden units in a single bi-directional GRU layer leads to the best result.

 \emph{Image Size}: We experimented with log-Mel spectra images of sizes $64 \times 64$ and $128 \times 128$ pixels and found that our model worked best with images of size of $128 \times 128$ pixels.

\vspace{5mm}
We also evaluated our model on data with mixup augmentation \cite{zhang2017mixup}. It is a data augmentation technique that also acts as a regularization technique and prevents overfitting. Instead of directly taking images from the training dataset as input, mixup takes a linear combination of any two random images and feeds it as input. The following equations were used to prepared a mixed-up dataset:

\begin{equation}
	\mathrm{Input\_Image} = \alpha * I_1 + (1 - \alpha) * I_2,
\end{equation}

and

\begin{equation}
	\mathrm{Input\_Label}	= \alpha * L_1 + (1 - \alpha) * L_2,
\end{equation}
where $\alpha \in [0, 1]$ is a random variable from a $\beta$-distribution, $I_1$.

\subsection{Model details: 2D-ConvNet}
\label{sec:2dcnn}
This model is a similar model to 2D-ConvNet with Attention and bi-directional GRU described in section~\ref{sec:2dgru} except that it lacks skip connections, attention layers, bi-directional GRU and the embedding layer incorporated in the previous model. 

\subsection{Dataset}
 
We classified six languages (English, French, German, Spanish, Russian and Italian) from the VoxForge \cite{VoxForge.org} dataset. VoxForge is an open-source speech corpus which primarily consists of samples recorded and submitted by users using their own microphone. This results in significant variation of speech quality between samples making it more representative of real world scenarios.

Our dataset consists of 1,500 samples for each of six languages. Out of 1,500 samples for each language, 1,200 were randomly selected as training dataset for that language and rest 300 as validation dataset using k-fold cross-validation. To sum up, we trained our model on 7,200 samples and validated it on 1800 samples comprising six languages. The results are discussed in next section. 

\section{Results and Discussion}
This paper discusses two end-to-end approaches which achieve state-of-the-art results in both the image as well as audio domain on the VoxForge dataset \cite{VoxForge.org}. In Table~\ref{tab:results}, we present all the classification accuracies of the two models of the cases with and without mixup for six and four languages.

\begin{figure}
 \begin{subfigure}{.5\textwidth}
 \centering\includegraphics[width=5cm]{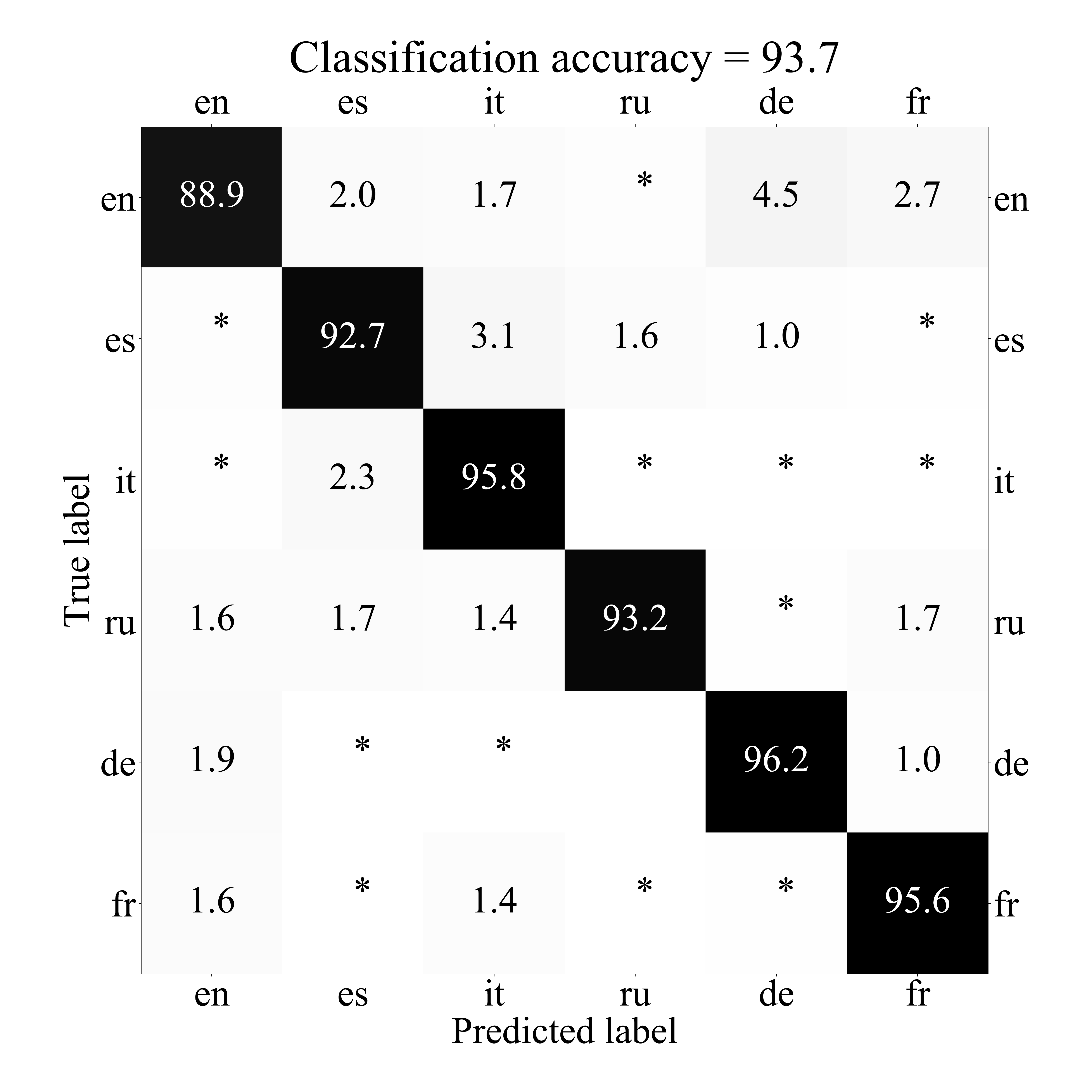}
 \caption{}
 \end{subfigure}
 \hspace{4pt}
 \begin{subfigure}{.5\textwidth}
 \centering\includegraphics[width=5cm]{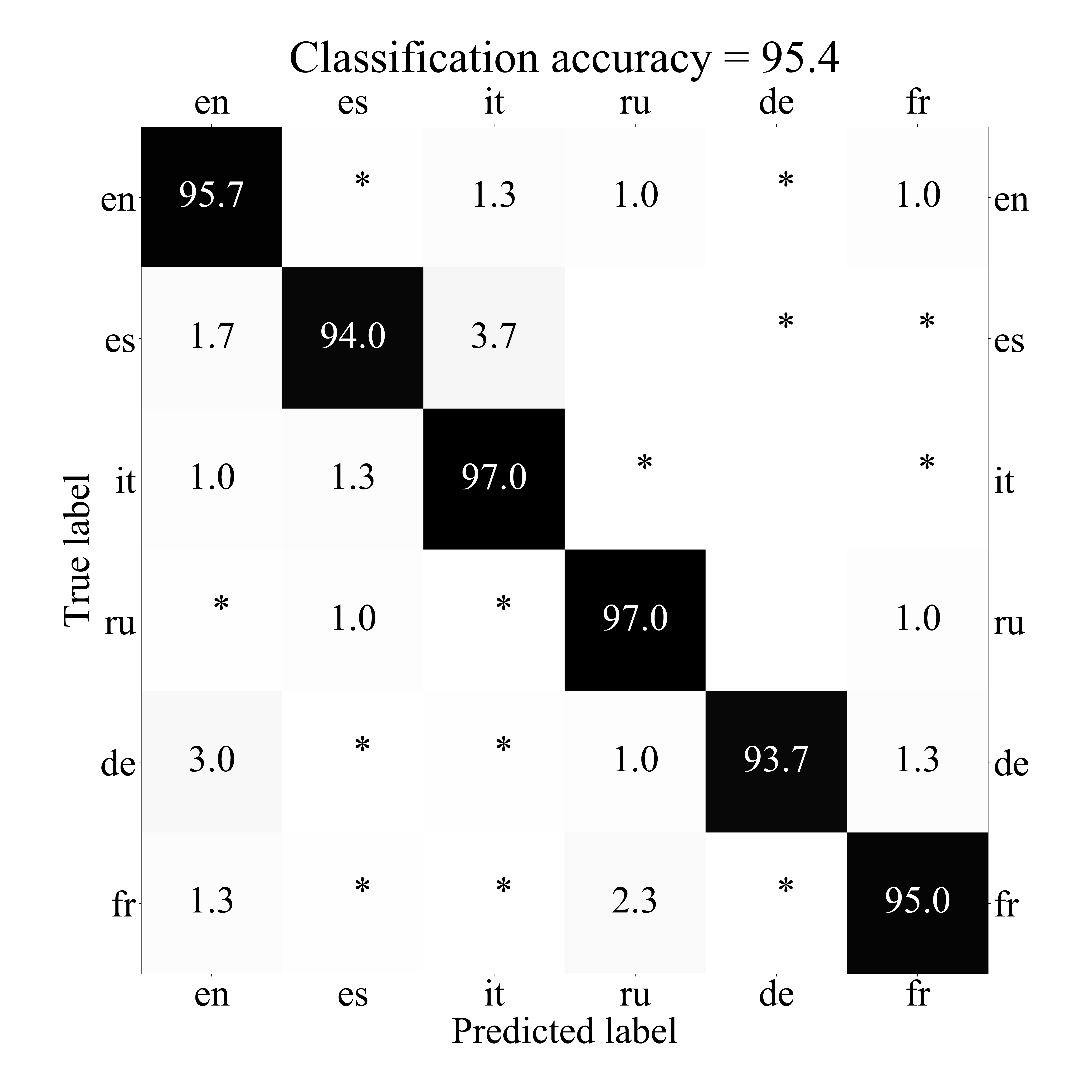}
 \caption{}
 \end{subfigure}
  \caption{Confusion matrix for classification of six languages with our (a) 1D-ConvNet and (b) 2D-ConvNet model. Asterisk (*) marks a value less than 0.1\%.}
 \label{fig:cnf_mat}
\vspace{-8mm}
\end{figure}

In the audio domain (using raw audio waveform as input), 1D-ConvNet achieved a mean accuracy of 93.7\% with a standard deviation of 0.3\% on running k-fold cross validation. In Fig~\ref{fig:cnf_mat} (a) we present the confusion matrix for the 1D-ConvNet model.

In the image domain (obtained by taking log-Mel spectra of raw audio), 2D-ConvNet with 2D attention (channel and spatial attention) and bi-directional GRU achieved a mean accuracy of 95.0\% with a standard deviation of 1.2\% for six languages. This model performed better when mixup regularization was applied. 2D-ConvNet achieved a mean accuracy of 95.4\% with standard deviation of 0.6\% on running k-fold cross validation for six languages when mixup was applied. In Fig~\ref{fig:cnf_mat} (b) we present the confusion matrix for the 2D-ConvNet model. 2D attention models focused on the important features extracted by convolutional layers and bi-directional GRU captured the temporal features. 

% results
\begin{table}[]
 \centering
 \setlength{\belowcaptionskip}{-10pt}
 \caption{Results of the two models and all its variations}
 \label{tab:results}

 \begin{tabular}{|l|l|l|l|l|}
 \hline
 \textbf{Languages} & \textbf{Feature Desc.} & \textbf{Network} & \textbf{Mixup} & \textbf{Accuracy} \\ \hline
 \multirow{5}{*}{\begin{tabular}[c]{@{}l@{}}En, Es, Fr, \\ De, Ru, It\end{tabular}} & Raw Waveform & 1D ConvNet & No & 93.7 \\ \cline{2-5} 
  & \multirow{4}{*}{log-Mel Spectra} & \multirow{2}{*}{2D ConvNet} & No & 94.3 \\ \cline{4-5} 
  &  &  & Yes & 95.4 \\ \cline{3-5} 
  &  & \multirow{2}{*}{\begin{tabular}[c]{@{}l@{}}2D ConvNet with\\ Attention and GRU\end{tabular}} & No & 94.3 \\ \cline{4-5} 
  &  &  & Yes & 95.0 \\ \hline
 \multirow{5}{*}{En, Es, Fr, De} & Raw Waveform & 1D ConvNet & No & 94.4 \\ \cline{2-5} 
  & \multirow{4}{*}{log-Mel Spectra} & \multirow{2}{*}{2D ConvNet} & No & 96.0 \\ \cline{4-5} 
  &  &  & Yes & 96.3 \\ \cline{3-5} 
  &  & \multirow{2}{*}{\begin{tabular}[c]{@{}l@{}}2D ConvNet with \\ Attention and GRU\end{tabular}} & No & 94.7 \\ \cline{4-5} 
  &  &  & Yes & 93.7 \\ \hline
 \end{tabular}%
\vspace{-5mm}
\end{table}

\subsubsection{Misclassification}
Several of the spoken languages in Europe belong to the Indo-European family. Within this family, the languages are divided into three phyla which are Romance, Germanic and Slavic. Of the 6 languages that we selected Spanish (Es), French (Fr) and Italian (It) belong to the Romance phyla, English and German belong to Germanic phyla and Russian in Slavic phyla. Our model also confuses between languages belonging to the similar phyla which acts as an insanity check since languages in same phyla have many similar pronounced words such as \emph{cat} in English becomes \emph{Katze} in German and \emph{Ciao} in Italian becomes \emph{Chao} in Spanish. 

Our model confuses between French (Fr) and Russian (Ru) while these languages belong to different phyla, many words from French were adopted into Russian such as automate (oot-oo-mate) in French becomes ABTOMaT (aff-taa-maat) in Russian which have similar pronunciation.

\vspace{-4mm}
% 1d convNet cnf_mat

\subsubsection{Future Scope}
The performance of raw audio waveforms as input features to ConvNet can be further improved by applying silence removal in the audio. Also, there is scope for improvement by augmenting available data through various conventional techniques like pitch shifting, adding random noise and changing speed of audio. These help in making neural networks more robust to variations which might be present in real world scenarios. There can be further exploration of various feature extraction techniques like Constant-Q transform and Fast Fourier Transform and assessment of their impact on Language Identification.

There can be further improvements in neural network architectures like concatenating the high level features obtained from 1D-ConvNet and 2D-ConvNet, before performing classification. There can be experiments using deeper networks with skip connections and Inception modules. These are known to have positively impacted the performance of Convolutional Neural Networks.

\section{Conclusion}

There are two main contributions of this paper in the domain of spoken language identification. Firstly, we presented an extensive analysis of raw audio waveforms as input features to 1D-ConvNet. We experimented with various hyperparameters in our 1D-ConvNet and evaluated their effect on validation accuracy. This method is able to bypass the computational overhead of conventional approaches which depend on generation of spectrograms as a necessary pre-procesing step. We were able to achieve an accauracy of \textbf{93.7\%} using this technique.

Next, we discussed the enhancement in performance of 2D-ConvNet using mixup augmentation, which is a recently developed technique to prevent overﬁtting on test data.This approach achieved an accuracy of \textbf{95.4\%}. We also analysed how \emph{attention} mechanism and recurrent layers impact the performance of networks. This approach achieved an accuracy of \textbf{95.0\%}.

\bibliographystyle{splncs04}
\bibliography{./bib/paper.bib}

\end{document}